\documentclass[letterpaper,twocolumn,10pt]{article}
\usepackage{usenix2019_v3}

\usepackage{amsmath,amssymb,amsfonts}
\usepackage{graphicx}
\usepackage{textcomp}
\usepackage{xcolor}
\usepackage{booktabs, multicol, multirow,bigdelim}
\usepackage[acronym,nomain]{glossaries}
\usepackage[inline]{enumitem}
\usepackage{algorithmicx}
\usepackage{romannum}
\usepackage{algpseudocode}
\usepackage{amsmath,amsfonts,amssymb}
\usepackage[hypcap=false]{caption}
\usepackage{subfigure}
\usepackage{grffile}
\usepackage{tabularx}
\usepackage{colortbl}
\usepackage{hhline}
\usepackage{enumerate}
\usepackage{balance}
\usepackage[flushleft]{threeparttable}
\usepackage{booktabs,fixltx2e}
\usepackage{balance}
\usepackage{boldline}
\usepackage{rotating}
\usepackage{pdflscape}
\usepackage[many]{tcolorbox}
\usepackage{framed}
\usepackage{color,soul}
\usepackage{flushend}
\usepackage{booktabs}
\usepackage{multirow}
\usepackage{siunitx}
\usepackage{romannum}
\usepackage{float}
\usepackage{array}
\usepackage[usestackEOL]{stackengine}[2013-10-15]
\usepackage[export]{adjustbox}
\usepackage{capt-of}
\usepackage{booktabs}
\usepackage{varwidth}
\usepackage{wrapfig}

\usepackage{empheq}

\definecolor{myblue}{rgb}{.8, .8, 1}

\newlength\mytemplendrcvv
\newsavebox\mytempbox

\makeatletter
\newcommand\mybluebox{%
    \@ifnextchar[
       {\@mybluebox}%
       {\@mybluebox[0pt]}}

\def\@mybluebox[#1]{%
    \@ifnextchar[
       {\@@mybluebox[#1]}%
       {\@@mybluebox[#1][0pt]}}

\def\@@mybluebox[#1][#2]#3{
    \sbox\mytempbox{#3}%
    \mytemplen\ht\mytempbox
    \advance\mytemplen #1\relax
    \ht\mytempbox\mytemplen
    \mytemplen\dp\mytempbox
    \advance\mytemplen #2\relax
    \dp\mytempbox\mytemplen
    \colorbox{myblue}{\hspace{1em}\usebox{\mytempbox}\hspace{1em}}}

\colorlet{shadecolor}{blue!20}

\renewcommand{\mathbf}[1]{{\boldsymbol #1}}
\setlist[itemize]{leftmargin=*}
\setlist[enumerate]{leftmargin=*}

\usepackage[many]{tcolorbox}
\usepackage{framed}
\usepackage{flushend}
\usepackage{enumitem}
\colorlet{shadecolor}{blue!20}
\usepackage{colortbl}

\usepackage{tikz}
\usepackage{amsmath}

\usepackage{filecontents}
\usepackage{titling}
\renewcommand{\mathbf}[1]{{\boldsymbol #1}}
\setlist[itemize]{leftmargin=*}
\setlist[enumerate]{leftmargin=*}

\begin{document}
\date{}


\title{\Large \bf Transfer Learning for Performance Modeling of Deep Neural Network Systems}
\author{
{\rm Md Shahriar Iqbal}\\
University of South Carolina
\and
{\rm Lars Kotthoff}\\
University of Wyoming
\and
{\rm Pooyan Jamshidi}\\
University of South Carolina
} 
\maketitle

\begin{abstract}
Modern deep neural network (DNN) systems are highly configurable with large a number of options that significantly affect their non-functional behavior, for example inference time and energy consumption. Performance models allow to understand and predict the effects of such configuration options on system behavior, but are costly to build because of large configuration spaces. Performance models from one environment cannot be transferred directly to another; usually models are rebuilt from scratch for different environments, for example different hardware. Recently, transfer learning methods have been applied to reuse knowledge from performance models trained in one environment in another. In this paper, we perform an empirical study to understand the effectiveness of different transfer learning strategies for building performance models of DNN systems. Our results show that transferring information on the most influential configuration options and their interactions is an effective way of reducing the cost to build performance models in new environments.
\end{abstract}
\vspace{-5mm}
\section{Introduction}
\vspace{-3mm}
\begin{wrapfigure}{r}{0.15\textwidth}

  \begin{center}
    \includegraphics[width=.3\textwidth,height=2.4 in,keepaspectratio=true]{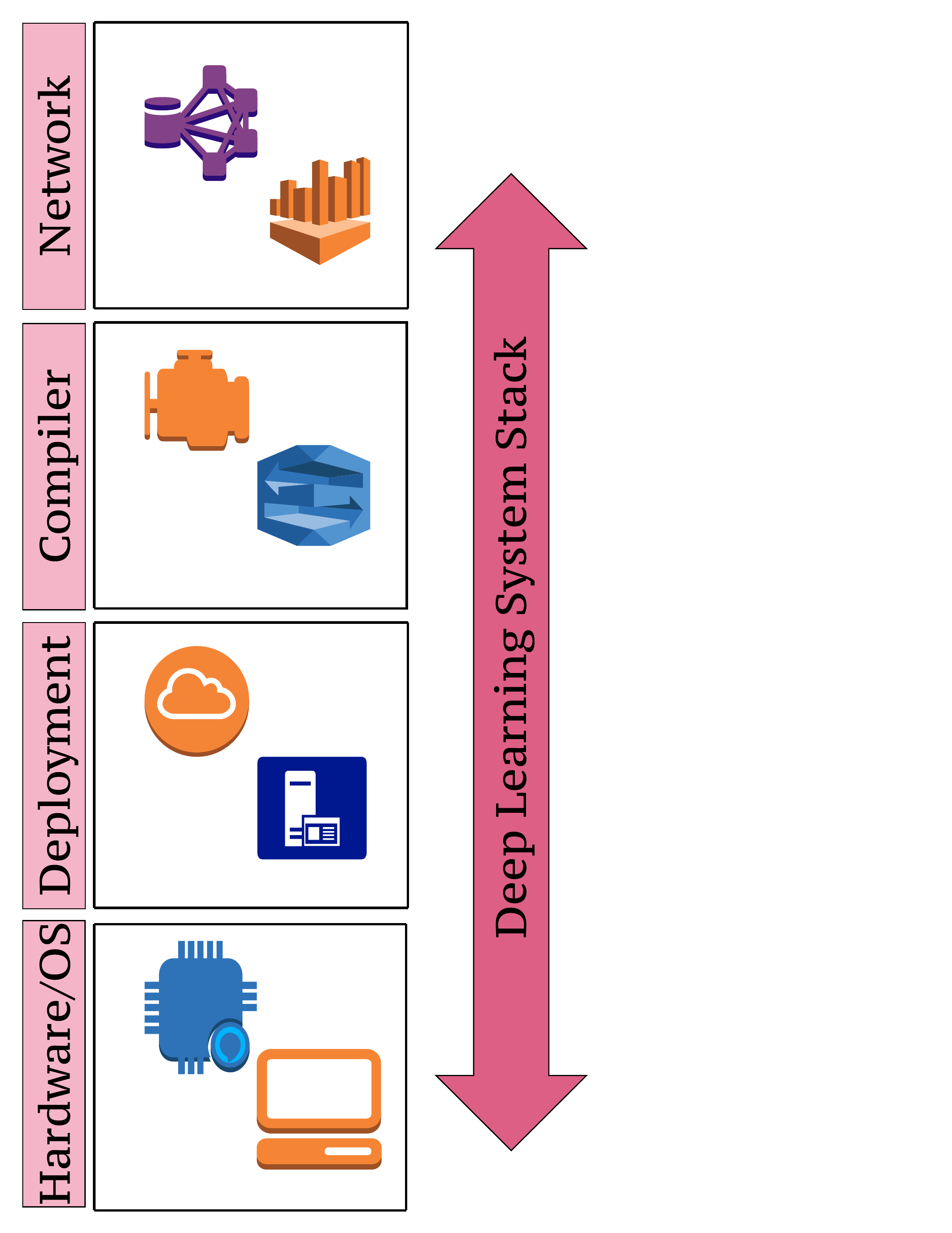}
  \end{center}
  \caption{DNN System Stack}
  \label{sys_stack}
\end{wrapfigure}
Deep neural networks (DNNs) are becoming increasingly complex, with an increased number of parameters to tune, and increased energy consumption for the deployed system \cite{sze2017efficient}. Current state-of-the-art methods for tuning DNNs do not consider how a DNN is deployed in a system stack \cite{cai2017neuralpower, sze2017efficient, chen2016eyeriss}, and do therefore not consider energy consumption. Figure~\ref{sys_stack} shows a \(4\)-level deployment environment of a DNN system where options and option interactions from each level contribute to energy consumption \cite{jamshidi2017transferseams, qi2016paleo, manotas2014seeds}.

Performance models have been extensively used for understanding the behavior of configurable systems \cite{GCASW:ASE13,HSCMAR:SIGPLAN,SGSAC:ASE15,SGAK:ESECFSE15,WWHJK:GECCO15,YWLE:MASCOTS13}. However, constructing such models requires extensive experimentation because of large parameter spaces, complex interactions, and unknown constraints \cite{XJFZPT:FSE15}. Such models are usually designed for fixed environments, i.e., fixed hardware and fixed workloads, and cannot be used directly when the environment changes. Repeating the process of building a performance model every time an environment change occurs is costly and slow. Several transfer learning approaches have been proposed to reuse information from performance models in a new environment for different configurable systems \cite{jamshidi2018learning,jamshidi2017transferseams,jamshidi2017transferase,geschwender_algorithm_2014}; however, to the best of our knowledge, no approach focuses specifically on DNNs in different environments.  
We consider the following research question:
\colorlet{shadecolor}{red!15}\vspace*{-2mm}
\begin{shaded*}  \noindent 
How can we efficiently and effectively transfer information from a performance model of a DNN trained for one environment to another environment?
\end{shaded*}
\vspace{-3mm}
We perform an empirical study comparing different transfer learning strategies for performance models of DNNs for different environmental changes, e.g., different hardware and different workloads. We consider guided sampling (GS)\cite{jamshidi2018learning}, direct model transfer (DM) \cite{VPGFC:ICPE17}, linear model shift (LMS), and non-linear model shift (NMLS)\cite{jamshidi2017transferseams}. We model the non-functional properties inference time and energy consumption in this paper and consider configuration options that affect these properties as the parameters we tune, i.e.\ hardware-level configuration options. Our results indicate that GS transfer learning outperforms next best learning method, NMLS, by \(19.76\%\) and \(8.33\%\) using regression trees (RT) and  by \(23.47\%\) and \(12.70\%\) using neural networks (NN) for inference time and energy consumption, respectively. This enabled us to build performance models in new environments using only \(2.44\%\) of the configuration space to predict best configurations in our systems with comparable accuracy to the performance models built for the original environment.The difference between the lowest and highest energy consumption can be up to a factor of 20.

\begin{figure*}[tb]
\centering
\subfigure{\includegraphics[width=\textwidth]{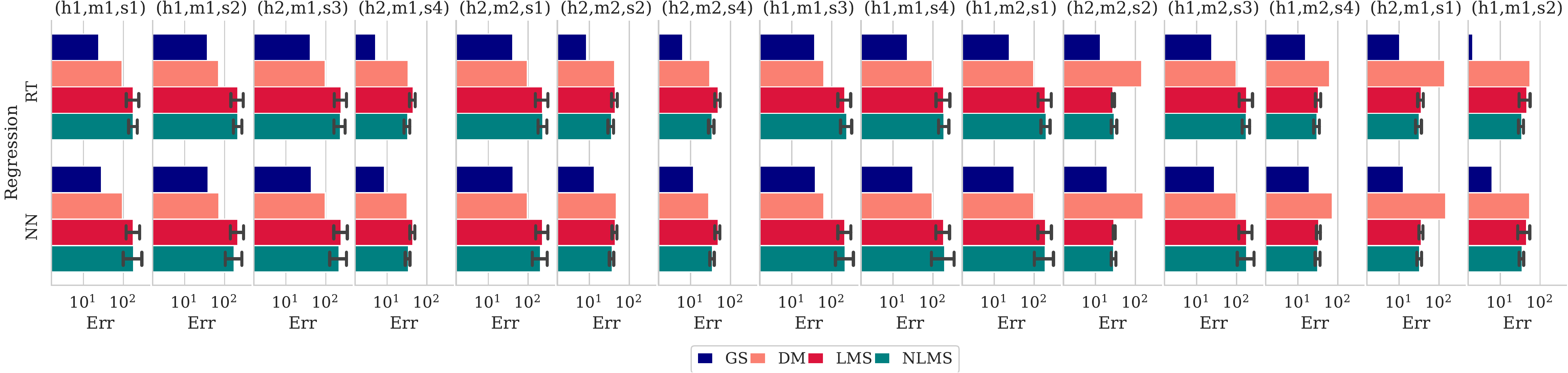}}
\subfigure{\includegraphics[width=\textwidth]{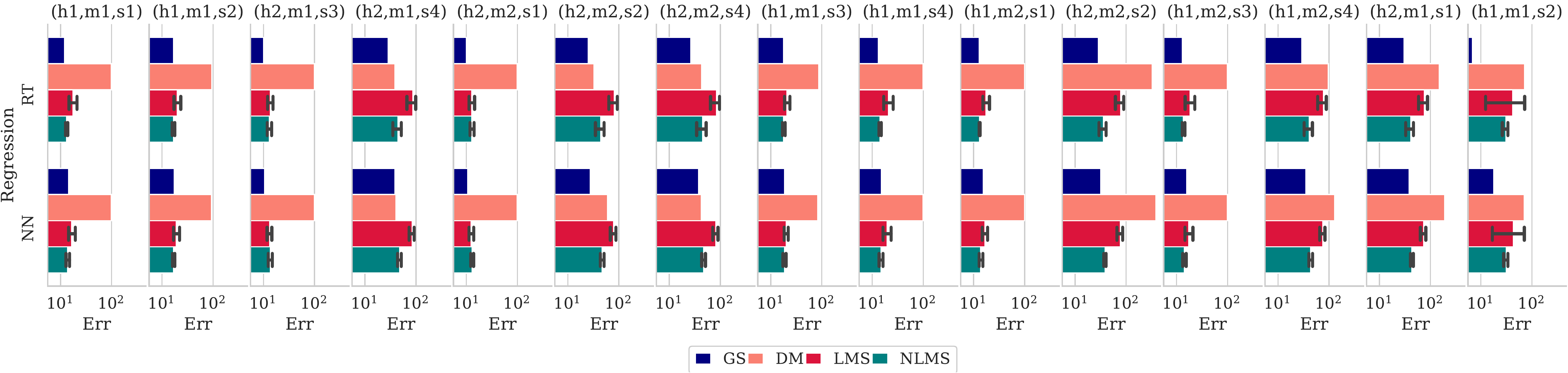}}
\caption{Comparison of prediction error of different transfer learning techniques (GS, DM, LMS, and NLMS) for performance models of DNN systems (Regression Tree and Neural Net) for inference time (top) and energy consumption (bottom). We consider 15 different target environments and show error bars for values aggregated over 10 predictions on a log scale.}
\label{res}

\end{figure*}

\vspace{-4mm}
\section{Methodology}
\vspace{-2mm}

We consider a pre-trained image recognition DNN system in 16 different environments: 2 different hardware platforms (Nvidia Jetson TX1, $h_1$, and Jetson TX2, $h_2$), 2 pre-trained models (Xception\cite{chollet2017xception}, $m_1$, and InceptionV3\cite{szegedy2016rethinking}, $m_2$) and 4 different image sizes ($200\times{}200$, $400\times{}400$, $600\times{}600$, and $800\times{}800$, $s_1$ through $s_4$). In each environment, we evaluate the performance on the same 10 randomly selected images from the ILSVRC2017\cite{ILSVRC15} image recognition dataset.

The configuration space we consider is composed of the following hardware configuration options: (\romannum{1}) CPU status, (\romannum{2}) CPU frequency, (\romannum{3}) GPU frequency, and (\romannum{4}) memory controller frequency. We evaluate a total of 46,080 configurations on the TX1 platform and 11,616 configurations on the TX2 platform, for a total experimental effort of $\approx$ 43.6 days of computational time across all 16 environments. We chose the TX1 and TX2 platforms due to their limited energy budget to better understand DNN system behavior with changing configuration options. 

We construct performance models of the effect of configuration options on DNN system performance using these experimental data with RTs and NNs, which are frequently used in the literature to induce performance models \cite{GCASW:ASE13,VPGFC:ICPE17,SGSAC:ASE15,NMSA:ASE,NMSA:FSE17}. We measure the performance of these models in terms of mean absolute percentage error, {\sf Err}\cite{MAPE}.

We implement \textbf{GS} using a step-wise regression technique with forward selection (FS) and backward elimination (BE). Each step of FS adds an interaction term to the regression model that increases the coefficient of determination, while BE removes an interaction term if its significance is below a threshold. We study the interaction terms of the final regression model; in particular, we exclude terms with coefficients that are less than \(10^{-12}\). These terms guide the sampling towards important configuration options and avoid wasting resources on evaluations that effect no change when building performance models in new environments. The \textbf{DM} transfer learning approach reuses a performance model built for one environment directly in a different environment. The \textbf{LMS} and \textbf{NMLS} transfer learning techniques learn a linear regression model and a non-linear random forest regression model, respectively, to translate the predictions from a performance model trained for one environment into predictions for a different environment. These transfer models are based on a small number of randomly-sampled configurations that are evaluated in both environments.

In our experiments, we select the TX2 platform with the InceptionV3 DNN and $600\times{}600$ images as the source environment to train the performance models for. We transfer these performance models to each of the remaining 15 target environments. The source code and data are available in an online appendix \cite{TL4DNN}. 
\vspace{-5mm}
\section{Results and Discussion}
\vspace{-3mm}
We present the results in Figure~\ref{res}. They demonstrate that GS outperforms DM, LMS, and NMLS in each environment for both inference time and energy consumption. Average {\sf Err} of the performance models induced using GS are \(28.09\%\) and \(22.93\%\) lower than DM, \(25.64\%\) and \(21.59\%\) lower than LMS, and \(23.47\%\) and \(19.76\%\) lower than NLMS for inference time using NN and RT, respectively. Similarly, they are \(42.85\%\) and \(39.41\%\) lower than DM, \(20.52\%\) and \(13.19\%\) lower than LMS, and \(12.70\%\) and \(8.33\%\) lower than NLMS for energy consumption for NN and RT, respectively. All of GS, LMS, and NLMS incurred the same cost (evaluation of \(2.44\%\) of the entire configuration space, \(\approx\) 2.48 hours), while the cost for DM was zero as the performance model from the source environment is reused without modification in the target environment. For the DM and GS transfer learning techniques, an increase in computational effort of just 2.48 hours ($\approx$ 0.15\% of the effort to train the original performance model) leads to an decrease of \({\sf Err}\) of  \(28.09\%\) and \(22.93\%\) for inference time and \(42.85\%\) and \(39.41\%\) for energy consumption using NN and RT, consecutively. 

If the environment change between source and target includes a hardware change, DM is more effective than LMS and NLMS for inference time modeling; however, for energy consumption, NLMS performs better than DM and LMS. 

Guided sampling can help practitioners to quickly develop reliable performance models for new environments based on information they have obtained in the past to tune and optimize a system. Such performance models can guide practitioners to avoid invalid configurations and are useful for design space exploration to quickly find optimal configurations in new environments using influential configurations which typically practitioners miss. These models are also useful to learn the performance landscape of a system for performance debugging, and obtain a better understanding of how the configuration options affect performance in general. In future work, we will consider extending the configuration space with options from all \(4\) levels of the DNN system stack.
\vspace{-4mm}
\section{Acknowledgements}
\vspace{-3mm}
This work has been supported by AFRL and DARPA (FA8750-16-2-0042). Lars Kotthoff is supported by NSF grant \#1813537.
\bibliographystyle{plain}
\bibliography{main}

\end{document}